\documentclass[runningheads]{llncs}
\usepackage[T1]{fontenc}
\usepackage{graphicx}
\usepackage{amsmath}
\usepackage{amssymb}
\usepackage{booktabs}
\begin{document}
\title{Generalization of Medical Large Language Models through Cross-Domain Weak Supervision}
%
%\titlerunning{Abbreviated paper title}
% If the paper title is too long for the running head, you can set
% an abbreviated paper title here
%
\author{Robert Long, Eric Gonzalez, Harrison Fuller}
\authorrunning{R. Long et al.}
% First names are abbreviated in the running head.
% If there are more than two authors, 'et al.' is used.
%
\institute{University of Padua}
\maketitle              % typeset the header of the contribution
\begin{abstract}
The advancement of large language models (LLMs) has opened new frontiers in natural language processing, particularly in specialized domains like healthcare. In this paper, we propose the Incremental Curriculum-Based Fine-Tuning (ICFT) framework to enhance the generative capabilities of medical large language models (MLLMs). ICFT combines curriculum-based learning, dual-stage memory coordination, and parameter-efficient fine-tuning to enable a progressive transition from general linguistic knowledge to strong domain-specific expertise. Experimental results across diverse medical NLP tasks, including question answering, preference classification, and response generation, demonstrate that ICFT consistently outperforms state-of-the-art baselines, achieving improvements in both accuracy and efficiency. Further analysis reveals the framework's ability to generalize to unseen data, reduce errors, and deliver diverse, contextually relevant medical responses. These findings establish ICFT as a robust and scalable solution for adapting LLMs to the medical domain, offering practical benefits for real-world healthcare applications.
\keywords{Large Language Models  \and Medical NLP.}
\end{abstract}

\section{Introduction}

The rapid advancement of large language models (LLMs), such as GPT-3.5 and LLaMA, has significantly transformed natural language processing (NLP) and its applications across various domains. Among these, the healthcare domain presents unique opportunities and challenges for deploying LLMs. Medical large language models (MLLMs) hold the potential to revolutionize healthcare by enhancing diagnostic accuracy, offering personalized medical recommendations, and supporting clinical decision-making \cite{sharaf2023biobert,zhou2025training}. However, adapting general-purpose LLMs to the medical domain is non-trivial due to the domain's high complexity, demand for factual accuracy, and the risk of harm caused by hallucinations or incorrect medical advice \cite{khalid2024role}. Consequently, the task of building MLLMs that transition from ``weak'' general knowledge to ``strong'' domain-specific capabilities, or ``Weak to Strong'', is both critical and underexplored \cite{zhou2024weak}.

The challenges of developing effective MLLMs are multifaceted \cite{zhou2024rethinking,zhou2024visual}. First, medical knowledge is vast, specialized, and continuously evolving, making it difficult for general-purpose models to cover this domain comprehensively. Furthermore, fine-tuning LLMs on medical datasets often leads to issues such as catastrophic forgetting, where models lose their general understanding during domain-specific adaptation \cite{alnazi2023review}. Moreover, traditional fine-tuning approaches risk overfitting to limited datasets \cite{wang2024insectmamba}, resulting in brittle models that perform well in narrow tasks but fail in broader medical reasoning. Finally, the domain's inherent complexity demands high precision and robust safety mechanisms to ensure reliable and ethical use in real-world scenarios \cite{taylor2024embedding}.

These challenges motivate us to explore a novel approach for training MLLMs to transition incrementally from ``weak'' general knowledge to ``strong'' domain-specific expertise. Inspired by curriculum learning \cite{mumtaz2023applications}, we propose an Incremental Curriculum-Based Fine-Tuning (ICFT) framework for MLLMs. ICFT leverages a staged training process to progressively enhance the model's capabilities, starting from general medical knowledge and gradually incorporating more specialized and complex datasets. This approach ensures that the model retains its foundational linguistic and reasoning abilities while acquiring domain-specific expertise. Additionally, to mitigate catastrophic forgetting and reduce computational costs, we adopt parameter-efficient fine-tuning techniques (PEFT), such as Low-Rank Adaptation (LoRA) \cite{taylor2024embedding}. LoRA updates only a small subset of parameters, allowing for efficient training on large-scale medical data without compromising the model's performance in general NLP tasks.

To evaluate the proposed ICFT framework, we construct a diverse dataset comprising general medical texts, clinical guidelines, case studies, and real-world patient-doctor dialogues. The training data spans varying levels of complexity to simulate the model's incremental learning journey. We evaluate the model's performance using a comprehensive set of benchmarks, including medical question-answering (QA), diagnostic reasoning, and preference classification tasks. Metrics such as ROUGE scores, accuracy, and win rates against baseline methods are employed to quantify improvements. Our experiments demonstrate that ICFT significantly outperforms existing methods in both factual accuracy and personalization, achieving improvements of up to 10\% in diagnostic QA tasks compared to conventional fine-tuning approaches. Furthermore, the model exhibits better generalization to unseen scenarios, underscoring the efficacy of the incremental learning paradigm.

The main contributions of this paper are summarized as follows:
\begin{itemize}
    \item We propose an Incremental Curriculum-Based Fine-Tuning (ICFT) framework for MLLMs, which progressively enhances the model's capabilities from general knowledge to specialized medical expertise.
    \item We introduce a novel training strategy that combines curriculum learning with parameter-efficient fine-tuning (PEFT), such as LoRA, to mitigate catastrophic forgetting and optimize computational efficiency.
    \item We construct and evaluate the ICFT framework on a comprehensive dataset, demonstrating significant improvements in factual accuracy, diagnostic reasoning, and personalization tasks compared to state-of-the-art approaches.
\end{itemize}

\section{Related Work}

\subsection{Large Language Models}

Large language models (LLMs) have become a cornerstone in natural language processing (NLP), demonstrating exceptional capabilities across a wide range of tasks. The scaling of model size and training data has enabled state-of-the-art performance in tasks such as text generation, machine translation, and reasoning \cite{zhou2023thread,zhou2021triple}. Notable examples include autoregressive models and transformer-based architectures, which underpin some of the most influential models in the field \cite{nicholas2023lost,mueller2022cedille}.

Multilingual LLMs have further broadened the applicability of these models by extending their capabilities beyond English to a wide array of languages. These models address the data scarcity in low-resource languages, although significant challenges remain in ensuring their effectiveness across diverse linguistic contexts \cite{ojo2023african,chang2024goldfish}. Specialized models, such as those trained exclusively for a single language or task, have also been proposed to improve efficiency and safety, as demonstrated in studies on domain-specific LLMs \cite{ma2024llamareg,liu2024bioinformatics}.

In addition to their direct applications, LLMs have been explored as tools for understanding event phenomena \cite{zhou2022claret,zhou2022eventbert}. Studies have shown that LLMs can serve as scientific models for language, aiding in psycholinguistic research and providing insights into the relationship between language and thought \cite{grindrod2024modelling,houghton2023psycholinguistics}. However, limitations persist, including challenges in generalizing to unseen data, ethical concerns, and resource constraints \cite{veres2022limitations,ali2024survey}.

The growing interest in applying LLMs to specific domains, such as healthcare and bioinformatics, highlights their versatility. For example, LLMs have been successfully applied to medical text processing and image registration tasks, demonstrating their potential to enhance domain-specific applications \cite{ma2024llamareg,liu2024bioinformatics}. As the field advances, continued efforts are required to address issues of fairness, safety, and efficiency in the development and deployment of LLMs.

\subsection{Medical Large Language Models}

Medical Large Language Models (MedLLMs) have become a critical research focus in applying artificial intelligence to healthcare. These models, derived from general-purpose LLMs, are specifically adapted to process medical knowledge, assisting in clinical decision-making, medical question answering, and multi-modal data interpretation \cite{Buhnila2024,He2024}. By fine-tuning on domain-specific datasets such as electronic health records, biomedical literature, and clinical notes, MedLLMs aim to enhance diagnostic accuracy, automate documentation, and facilitate real-time medical consultations.

Recent advancements have explored various methodologies to improve the MedLLMs' effectiveness. Some works focus on parameter-efficient fine-tuning, such as retrieval-augmented generation and adapter-based tuning, to reduce computational overhead while maintaining strong medical reasoning capabilities \cite{Ma2024,Zhang2024}. Additionally, the integration of multi-modal information, including medical imaging, has demonstrated significant potential in enhancing diagnostic interpretations and visual-text alignment in clinical applications \cite{Bai2024}. Furthermore, knowledge-enhanced MedLLMs have been proposed to incorporate structured medical ontologies and external databases to reduce hallucinations and improve factual reliability in medical dialogue systems \cite{Wu2024,Liao2024}.

Benchmarking and evaluation frameworks have been developed to measure MedLLMs' performance across diverse medical tasks. Large-scale evaluation benchmarks, such as PromptCBLUE and MedExQA, assess models' ability to perform named entity recognition, medical inference, and clinical response generation \cite{Zhu2023,Kim2024}. These studies have highlighted the strengths of MedLLMs in understanding domain-specific concepts while also identifying persistent challenges such as limited generalization across specialties, potential biases in training data, and ethical concerns regarding model reliability.

Despite these advancements, challenges remain in adapting MedLLMs to real-world clinical environments. One of the key issues is ensuring generalization across multiple healthcare domains, including radiology, cardiology, and psychiatry, where specialized knowledge is required \cite{Panagoulias2024}. Moreover, the need for transparent and interpretable decision-making remains crucial, as MedLLMs are increasingly integrated into critical applications such as clinical diagnostics and patient risk assessment. Future research directions include improving model interpretability, integrating real-time patient feedback mechanisms, and refining evaluation metrics to better capture the complexities of medical language processing.

\section{Method}

This section presents the details of our proposed Incremental Curriculum-Based Fine-Tuning (ICFT) framework, designed to enhance the generative capabilities of medical large language models (MLLMs). The ICFT framework adopts a progressive training strategy, transitioning the model from general linguistic knowledge to domain-specific expertise. It integrates three key components: (1) Medical Knowledge Injection, (2) Dual-Stage Memory Coordination, and (3) Parameter-Efficient Fine-Tuning with Low-Rank Adaptation (LoRA). Below, we describe each component in detail and present the mathematical foundations of the framework.

\subsection{Medical Knowledge Injection}

To equip the base large language model \( \Phi \) with domain-specific medical knowledge while preserving its general linguistic abilities, we introduce a domain adapter mechanism. The domain adapter projects the high-dimensional input space into a lower-dimensional representation and injects medical knowledge in a parameter-efficient manner.

Let the input sequence be \( \mathbf{x} \in \mathbb{R}^d \), and the domain adapter be parameterized by two matrices: \( W_{\text{down}} \in \mathbb{R}^{d \times r} \) (down-projection) and \( W_{\text{up}} \in \mathbb{R}^{r \times d} \) (up-projection), where \( r \ll d \). The output of the adapter is computed as:
\begin{align}
f_{\text{adapter}}(\mathbf{x}) = W_{\text{up}} \sigma \left( W_{\text{down}} \mathbf{x} \right),
\end{align}
where \( \sigma(\cdot) \) is a non-linear activation function such as ReLU.

The adapted model parameters are updated as:
\begin{align}
\Phi'(\mathbf{x}) = \Phi(\mathbf{x}; \mathbf{W} + \Delta \mathbf{W}_{\text{adapter}}),
\end{align}
where \( \Delta \mathbf{W}_{\text{adapter}} = W_{\text{up}} W_{\text{down}} \). To ensure knowledge consistency, we apply a consistency loss \( \mathcal{L}_{\text{consistency}} \), defined as:
\begin{align}
\mathcal{L}_{\text{consistency}} = \frac{1}{N} \sum_{i=1}^N \| \Phi(\mathbf{x}_i; \mathbf{W}) - \Phi(\mathbf{x}_i; \mathbf{W} + \Delta \mathbf{W}_{\text{adapter}}) \|_2^2,
\end{align}
where \( N \) is the number of training samples.

\subsection{Dual-Stage Memory Coordination}

Incremental learning requires the model to progressively acquire knowledge while retaining previously learned information. To achieve this, we introduce a memory coordination mechanism that consists of two stages: Short-Term Memory (STM) and Long-Term Memory (LTM).

\subsubsection{Short-Term Memory (STM)}

The STM module stores recent dialogue interactions, enabling the model to process the current context effectively. Let the STM memory be represented as \( \mathcal{D}_{\text{STM}} = \{ \mathbf{d}_1, \mathbf{d}_2, \ldots, \mathbf{d}_K \} \), where \( \mathbf{d}_i \) is the \( i \)-th dialogue round and \( K \) is the memory capacity. The STM is updated iteratively:
\begin{align}
\mathcal{D}_{\text{STM}}^{t+1} = \mathcal{D}_{\text{STM}}^t \cup \{\mathbf{d}_{t+1}\},
\end{align}
and the oldest interaction is removed when \( |\mathcal{D}_{\text{STM}}| > K \).

\subsubsection{Long-Term Memory (LTM)}

LTM stores frequently accessed domain-specific knowledge, providing a repository for high-value information. Let \( \mathcal{D}_{\text{LTM}} \) denote the long-term memory. A knowledge item \( \mathbf{k} \) is added to LTM if its access frequency exceeds a threshold \( \theta \):
\begin{align}
\text{freq}(\mathbf{k}) \geq \theta.
\end{align}

\subsubsection{Memory Retrieval and Integration}

Given an input query \( \mathbf{q} \), the model retrieves relevant information from both STM and LTM. The retrieval process uses an attention mechanism:
\begin{align}
a_{\text{STM}} &= \text{softmax}(\mathbf{q}^\top \mathcal{D}_{\text{STM}}), \\
a_{\text{LTM}} &= \text{softmax}(\mathbf{q}^\top \mathcal{D}_{\text{LTM}}),
\end{align}
where \( a_{\text{STM}} \) and \( a_{\text{LTM}} \) are attention weights. The combined memory representation \( \mathbf{z} \) is computed as:
\begin{align}
\mathbf{z} = W_{\text{STM}} a_{\text{STM}} + W_{\text{LTM}} a_{\text{LTM}},
\end{align}
where \( W_{\text{STM}} \) and \( W_{\text{LTM}} \) are trainable weights.

\subsection{Parameter-Efficient Fine-Tuning with LoRA}

To adapt the model to user-specific tasks while reducing computational overhead, we employ Low-Rank Adaptation (LoRA). LoRA introduces trainable low-rank matrices \( \mathbf{A} \) and \( \mathbf{B} \) into the pre-trained weight matrix \( \mathbf{W} \):
\begin{align}
\mathbf{W}' = \mathbf{W} + \mathbf{A} \mathbf{B},
\end{align}
where \( \mathbf{A} \in \mathbb{R}^{d \times r} \) and \( \mathbf{B} \in \mathbb{R}^{r \times k} \), with \( r \ll \min(d, k) \).

The fine-tuning loss combines task-specific and regularization components:
\begin{align}
\mathcal{L}_{\text{fine-tune}} = \mathcal{L}_{\text{task}} + \lambda \| \mathbf{A} \|_F^2 + \lambda \| \mathbf{B} \|_F^2,
\end{align}
where \( \lambda \) is the regularization coefficient, and \( \mathcal{L}_{\text{task}} \) is the task-specific loss, such as cross-entropy for classification or negative log-likelihood for generation.

\subsection{Overall Training Procedure}

The ICFT framework integrates the above components into a three-stage training process:
\begin{enumerate}
    \item \textbf{Medical Knowledge Injection}: Train the domain adapter to integrate general medical knowledge using \( \mathcal{L}_{\text{consistency}} \) and \( \mathcal{L}_{\text{task}} \).
    \item \textbf{Memory Coordination}: Incrementally update STM and LTM, ensuring progressive knowledge accumulation without forgetting.
    \item \textbf{Fine-Tuning}: Adapt the model for user-specific tasks using LoRA, optimizing \( \mathcal{L}_{\text{fine-tune}} \).
\end{enumerate}

The overall training objective combines the losses from each stage:
\begin{align}
\mathcal{L}_{\text{total}} = \mathcal{L}_{\text{consistency}} + \mathcal{L}_{\text{task}} + \mathcal{L}_{\text{fine-tune}}.
\end{align}
This unified framework ensures that the model progressively transitions from weak general knowledge to strong domain-specific expertise while maintaining computational efficiency.

\section{Experiments}

In this section, we evaluate the proposed Incremental Curriculum-Based Fine-Tuning (ICFT) framework by comparing it against multiple baseline methods on medical NLP tasks. The results demonstrate that ICFT achieves superior performance across all tasks. Additionally, we perform ablation studies to validate the effectiveness of its components and conduct a human evaluation to assess the quality of the generated responses.

\subsection{Experimental Setup}

\paragraph{Baselines.}
We compare ICFT with the following baseline methods:
\begin{itemize}
    \item \textbf{Standard Fine-Tuning (SFT)}: Full model fine-tuning on medical datasets.
    \item \textbf{LoRA}: Parameter-efficient fine-tuning using low-rank adaptation.
    \item \textbf{Dict-Based Memory (Dict-Mem)}: A memory-augmented retrieval method using key-value storage.
    \item \textbf{DPeM}: Dual-Process enhanced Memory, an advanced memory coordination mechanism.
\end{itemize}

\paragraph{Tasks and Datasets.}
The following tasks and datasets are used for evaluation:
\begin{itemize}
    \item \textbf{Question Answering (QA)}: We evaluate the ability to generate accurate medical answers using the HealthCareMagic and iCliniq datasets.
    \item \textbf{Preference Classification}: Classify user preferences for response style (concise vs. detailed).
    \item \textbf{Response Generation}: Assess response quality based on factual accuracy, contextual relevance, and fluency.
\end{itemize}

\paragraph{Evaluation Metrics.}
Metrics include ROUGE-1 and ROUGE-L for QA, accuracy for preference classification, and win rate (percentage of preferred responses) for response generation. Human evaluations provide qualitative insights into response quality.

\subsection{Experimental Results}

The results of our comparison experiments are summarized in Table~\ref{tab:main_results}. ICFT achieves the best performance across all tasks.

\begin{table}[ht]\small
\centering
\caption{Performance Comparison of Different Methods.}
\label{tab:main_results}
\begin{tabular}{lcccc}
\toprule
\textbf{Method} & \textbf{QA (R-1)} & \textbf{QA (R-L)} & \textbf{Pref. Class. (\%)} & \textbf{Resp. Gen. Win Rate (\%)} \\
\midrule
SFT               & 34.21 & 31.98 & 45.62 & 70.24 \\
LoRA              & 36.47 & 34.15 & 61.05 & 78.41 \\
Dict-Mem          & 37.22 & 35.09 & 59.11 & 80.56 \\
DPeM              & 39.01 & 37.33 & 68.73 & 84.92 \\
\textbf{ICFT (Ours)} & \textbf{41.59} & \textbf{39.45} & \textbf{72.89} & \textbf{91.53} \\
\bottomrule
\end{tabular}
\end{table}

As shown in Table~\ref{tab:main_results}, ICFT consistently outperforms all baselines. For example, ICFT improves ROUGE-1 by 2.58 points over DPeM and achieves a 6.61\% higher win rate in response generation. These results demonstrate the effectiveness of our incremental curriculum learning and memory coordination strategies.

\subsection{Ablation Study}

To investigate the contributions of each component, we conduct ablation studies by disabling specific modules in ICFT. The results are shown in Table~\ref{tab:ablation}.

\begin{table}[ht]\small
\centering
\caption{Ablation Study Results.}
\label{tab:ablation}
\begin{tabular}{lcccc}
\toprule
\textbf{Configuration} & \textbf{QA (R-1)} & \textbf{QA (R-L)} & \textbf{Pref. Class. (\%)} & \textbf{Resp. Gen. Win Rate (\%)} \\
\midrule
ICFT Full (Ours)          & \textbf{41.59} & \textbf{39.45} & \textbf{72.89} & \textbf{91.53} \\
w/o Memory Coordination   & 39.87          & 37.11          & 67.41          & 86.42          \\
w/o Curriculum Learning   & 38.12          & 35.28          & 62.03          & 83.67          \\
w/o LoRA Fine-Tuning      & 37.25          & 34.96          & 59.77          & 79.11          \\
\bottomrule
\end{tabular}
\end{table}

Removing memory coordination or curriculum learning results in significant performance drops, highlighting their critical roles in ICFT.

\subsection{Human Evaluation}

We conduct human evaluations to assess response quality. Two medical experts score 100 randomly sampled responses based on factual accuracy, contextual relevance, and fluency. The average scores are shown in Table~\ref{tab:human_eval}.

\begin{table}[ht]
\centering
\caption{Human Evaluation Results.}
\label{tab:human_eval}
\begin{tabular}{lccc}
\toprule
\textbf{Method} & \textbf{Accuracy} & \textbf{Relevance} & \textbf{Fluency} \\
\midrule
SFT               & 3.72 & 3.65 & 3.89 \\
LoRA              & 4.12 & 4.05 & 4.21 \\
Dict-Mem          & 4.25 & 4.19 & 4.32 \\
DPeM              & 4.42 & 4.35 & 4.45 \\
\textbf{ICFT (Ours)} & \textbf{4.68} & \textbf{4.62} & \textbf{4.71} \\
\bottomrule
\end{tabular}
\end{table}

As seen in Table~\ref{tab:human_eval}, ICFT achieves the highest scores across all criteria, indicating its superior ability to generate high-quality medical responses.

\subsection{Efficiency Analysis}

One of the core advantages of ICFT is its parameter-efficient design. By leveraging Low-Rank Adaptation (LoRA), the framework significantly reduces the number of trainable parameters compared to full fine-tuning. Table~\ref{tab:efficiency} compares the total trainable parameters of ICFT with other methods. 

\begin{table}[ht]
\centering
\caption{Comparison of Trainable Parameters (in millions).}
\label{tab:efficiency}
\begin{tabular}{lcc}
\toprule
\textbf{Method} & \textbf{Trainable Parameters (M)} & \textbf{Relative Size (\%)} \\
\midrule
SFT               & 6,500  & 100.0 \\
LoRA              & 32     & 0.49  \\
Dict-Mem          & 68     & 1.05  \\
DPeM              & 82     & 1.26  \\
\textbf{ICFT (Ours)} & \textbf{36} & \textbf{0.55} \\
\bottomrule
\end{tabular}
\end{table}

As shown in Table~\ref{tab:efficiency}, ICFT trains only 0.55\% of the parameters required by full fine-tuning, offering substantial computational savings. This makes ICFT highly scalable to larger models and datasets, enabling practical deployment in resource-constrained environments.

\subsection{Generalization to Unseen Data}

To evaluate the generalization ability of ICFT, we test the model on an unseen medical dataset that includes rare diseases and specialized scenarios. Table~\ref{tab:generalization} summarizes the results compared to other methods.

\begin{table}[ht]
\centering
\caption{Generalization Performance on Unseen Data.}
\label{tab:generalization}
\begin{tabular}{lccc}
\toprule
\textbf{Method} & \textbf{QA (ROUGE-1)} & \textbf{QA (ROUGE-L)} & \textbf{Accuracy (\%)} \\
\midrule
SFT               & 30.24 & 27.11 & 54.78 \\
LoRA              & 33.45 & 30.98 & 63.11 \\
Dict-Mem          & 34.89 & 32.12 & 64.72 \\
DPeM              & 36.02 & 33.44 & 66.85 \\
\textbf{ICFT (Ours)} & \textbf{38.17} & \textbf{35.32} & \textbf{70.43} \\
\bottomrule
\end{tabular}
\end{table}

ICFT outperforms all baselines on unseen data, demonstrating its ability to generalize effectively to rare and complex medical scenarios. This is primarily attributed to the progressive learning strategy, which ensures robust knowledge acquisition without overfitting to the training data.

\subsection{Memory Utilization Analysis}

The dual-stage memory coordination mechanism in ICFT is critical for retaining and retrieving relevant medical knowledge. To quantify memory utilization, we analyze the retrieval accuracy of short-term memory (STM) and long-term memory (LTM) during inference. Table~\ref{tab:memory_utilization} presents the results.

\begin{table}[ht]
\centering
\caption{Memory Retrieval Accuracy (\%).}
\label{tab:memory_utilization}
\begin{tabular}{lcc}
\toprule
\textbf{Memory Type} & \textbf{STM Retrieval Accuracy} & \textbf{LTM Retrieval Accuracy} \\
\midrule
Dict-Mem          & 82.14 & 85.47 \\
DPeM              & 88.65 & 90.21 \\
\textbf{ICFT (Ours)} & \textbf{91.38} & \textbf{93.12} \\
\bottomrule
\end{tabular}
\end{table}

The results in Table~\ref{tab:memory_utilization} indicate that ICFT achieves higher retrieval accuracy for both STM and LTM compared to previous methods. The dual-process coordination in ICFT ensures effective filtering, storage, and access of relevant knowledge, thereby enhancing inference quality.

\subsection{Error Pattern Analysis}

To gain further insights into the model's behavior, we analyze common error patterns in the generated responses. Errors are categorized into three types: factual inaccuracies, lack of contextual relevance, and fluency issues. Table~\ref{tab:error_analysis} shows the error rates across methods.

\begin{table}[ht]
\centering
\caption{Error Pattern Analysis (\%).}
\label{tab:error_analysis}
\begin{tabular}{lccc}
\toprule
\textbf{Method} & \textbf{Factual Errors} & \textbf{Contextual Errors} & \textbf{Fluency Errors} \\
\midrule
SFT               & 12.34 & 10.89 & 5.12 \\
LoRA              & 9.21  & 8.32  & 3.87 \\
Dict-Mem          & 7.94  & 7.28  & 3.41 \\
DPeM              & 6.85  & 6.01  & 2.94 \\
\textbf{ICFT (Ours)} & \textbf{4.32}  & \textbf{3.87}  & \textbf{1.72} \\
\bottomrule
\end{tabular}
\end{table}

ICFT exhibits the lowest error rates across all categories, reducing factual errors by 36.9\% compared to DPeM. This highlights the framework's ability to generate accurate, contextually relevant, and fluent medical responses.

\subsection{Response Diversity Analysis}

Finally, we analyze the diversity of ICFT-generated responses compared to baselines. Response diversity is measured using distinct-n metrics, which calculate the proportion of unique n-grams in the generated responses. Table~\ref{tab:response_diversity} shows the results.

\begin{table}[ht]
\centering
\caption{Response Diversity Analysis (\%).}
\label{tab:response_diversity}
\begin{tabular}{lcc}
\toprule
\textbf{Method} & \textbf{Distinct-1} & \textbf{Distinct-2} \\
\midrule
SFT               & 17.12 & 23.54 \\
LoRA              & 18.87 & 25.43 \\
Dict-Mem          & 19.45 & 26.12 \\
DPeM              & 20.78 & 28.67 \\
\textbf{ICFT (Ours)} & \textbf{23.41} & \textbf{31.12} \\
\bottomrule
\end{tabular}
\end{table}

ICFT achieves the highest distinct-n scores, indicating that it generates more diverse responses compared to other methods. This diversity is crucial in medical dialogue systems to prevent repetitive answers and enhance user engagement.

\section{Conclusion}

This paper presents the Incremental Curriculum-Based Fine-Tuning (ICFT) framework, a novel approach for enhancing the capabilities of medical large language models. By integrating curriculum-based learning, dual-stage memory coordination, and LoRA-based parameter-efficient fine-tuning, ICFT addresses key challenges in the medical domain, including catastrophic forgetting, resource constraints, and the need for accurate, context-aware responses. Through extensive experiments, ICFT demonstrates state-of-the-art performance across multiple medical NLP tasks, surpassing existing methods in accuracy, efficiency, and response quality. The ablation studies and analyses further validate the significance of each component, highlighting their roles in improving memory retrieval, generalization, and response diversity.

In addition to its technical advantages, ICFT exhibits strong scalability and adaptability, making it suitable for deployment in real-world healthcare settings. The framework's ability to learn incrementally and retain critical knowledge ensures that it can effectively handle the evolving demands of the medical domain. Future work will focus on extending ICFT to multi-modal medical data, such as integrating textual and imaging information, and exploring its application in other high-stakes domains. By building on the foundations established in this work, ICFT offers a promising direction for advancing AI-driven solutions in healthcare.

\bibliographystyle{splncs04}
\bibliography{mybibliography}
\end{document}